\documentclass[pmlr]{jmlr}

\usepackage{longtable}

\usepackage{booktabs}
\usepackage[load-configurations=version-1]{siunitx} 

\makeatletter
\def\set@curr@file#1{\def\@curr@file{#1}} 
\makeatother

\theorembodyfont{\upshape}
\theoremheaderfont{\scshape}
\theorempostheader{:}
\theoremsep{\newline}

\newcommand{\tb}{\textbf}
\usepackage{enumitem}
\usepackage{multirow}

\title[On the Generation of Medical Dialogues for COVID-19]{On the Generation of Medical Dialogues for COVID-19}

\author{\Name{Guangtao Zeng\textsuperscript{1*$\dagger$}}, \Name{Wenmian Yang\textsuperscript{1*}}, \Name{Bowen Tan\textsuperscript{2*}},  \Name{Zeqian Ju\textsuperscript{1$\dagger$}}, \Name{Subrato Chakravorty\textsuperscript{1}}, \Name{Xuehai He\textsuperscript{1}}, \Name{Shu Chen\textsuperscript{1$\dagger$}}, \Name{Xingyi Yang\textsuperscript{1}}, \Name{Qingyang Wu\textsuperscript{3}}, \Name{Zhou Yu\textsuperscript{3}}, \Name{Eric Xing\textsuperscript{2}}, \Name{Pengtao Xie\textsuperscript{1}}\\
\addr UC San Diego\textsuperscript{1}, CMU\textsuperscript{2}, UC Davis\textsuperscript{3} \\
\Email{pengtaoxie2008@gmail.com}
       }

\begin{document}

\maketitle

\begin{abstract}
Under the pandemic of COVID-19, people experiencing COVID19-related symptoms or exposed to risk factors have a pressing need to consult doctors. Due to hospital closure, a lot of consulting services have been moved online. Because of the shortage of medical professionals, many people cannot receive online consultations timely. To address this problem, we aim to develop a medical dialogue system that can provide COVID19-related consultations. We collected two dialogue datasets -- CovidDialog -- (in English and Chinese respectively) containing conversations between doctors and patients about COVID-19. On these two datasets, we train several dialogue generation models based on Transformer, GPT, and BERT-GPT. Since the two COVID-19 dialogue datasets are small in size, which bear high risk of overfitting, we leverage transfer learning to mitigate data deficiency. Specifically, we take the pretrained models of Transformer, GPT, and BERT-GPT on dialog datasets and other large-scale texts, then finetune them on our CovidDialog tasks. We perform both automatic and human evaluation of responses generated by these models. The results show that the generated responses are promising in being doctor-like, relevant to the conversation history, and clinically informative.  The data and code are available at \url{https://github.com/UCSD-AI4H/COVID-Dialogue} 
\end{abstract}

\section{Introduction}

\footnotetext{*Equal contribution}
\footnotetext{$\dagger$The work was done during internship at UCSD.}

As of June 3rd in 2020, the COVID-19 pandemic has killed 386,581 people out of 6,542,851 infected cases. People who are experiencing symptoms (e.g., fever, cough) similar to those of COVID-19 or were exposed to risk factors such as close contact with infected cases have a pressing need to consult doctors, largely because of the panic over this unknown new disease. However, under the pandemic situation, coming to hospitals is dangerous and has high risk of suffering cross-infection. Cross-infection refers to the fact that many people visiting hospitals at the same time and infected individuals will spread coronavirus to healthy ones. To prevent spreading of the coronavirus, many non-urgent clinics and hospitals have been closed physically and encourage people to consult doctors through telemedicine services (e.g., phone calls, video conferencing). However, medical professionals are highly occupied by taking care of the infected patients and have very thin bandwidth to deal with the surging requests of consultations related to COVID-19. As a result, many people could not receive timely advice for effectively dealing with their medical conditions.

To address the large imbalance between the surging need of consultations from citizens and the severe shortage of medical professionals available to provide online consultation services, it is highly valuable to develop intelligent dialogue systems which act as “virtual doctors” to provide COVID-related consultations to people. These “virtual doctors” can greatly ease the burden of human doctors and timely address the concerns of the public. 


To facilitate the research and development of COVID19-targeted dialogue systems, we build two medical dialogue datasets that contain conversations between doctors and patients, about COVID-19 and other pneumonia: (1) an English dataset containing 603 consultations, 1232 utterances, and 90664 tokens (English words); (2) a Chinese dataset containing 1088 consultations, 9494 utterances, and 406550 tokens (Chinese characters). 

On these two datasets, we train several dialogue generation models based on Transformer~\citep{vaswani2017attention}, GPT~\citep{radford2018improving,zhang2019dialogpt}, and BERT-GPT~\citep{wu2019importance,lewis2019bart}. Transformer is an encoder and decoder architecture which takes  conversation history as inputs and generates response. Self-attention is used to capture long-range dependency among tokens. GPT is a language model based on the Transformer decoder. When generating a response, GPT predicts the next token using its context including the already decoded tokens in this response and the conversation history. BERT-GPT is an encoder-decoder architecture as well where the pretrained BERT~\citep{devlin2018bert} is used to encode the conversation history and GPT is used to decode the response. The small size of CovidDialog datasets incurs high risk of overfitting, if directly training the large-sized neural models on CovidDialog. To alleviate this risk, we take the pretrained weights of these models on large-scale dialogue datasets and other corpus and finetune the weights on CovidDialog. We perform automatic evaluation and human evaluation. The results show that the generated responses demonstrate high potential to be doctor-like, relevant to patient history, and clinically informative, which paves the way for building a COVID-19 consultation chatbot.  


The major contributions of this paper are as follows:
\begin{itemize}[leftmargin=*]
    \item We collect two medical dialogue datasets about COVID-19: one in English, the other in Chinese.
     \item We train several dialogue generation models on the collected datasets.
      \item We perform human evaluation and automatic evaluation of the generated responses. The results show that the generated responses are promising in being doctor-like, relevant, and informative. 
\end{itemize}

The rest of the paper is organized as follows. Section 2 and 3 present  datasets and methods. Section 4 gives experimental results. Section 5 reviews related works and Section 6 concludes the paper.

\section{Dataset}

In this section, we present two collected datasets -- CovidDialog-English and CovidDialog-Chinese -- which contain medical conversations between patients and doctors about COVID-19 and other related pneumonia. 
The statistics of these two datasets are summarized in Table~\ref{tb:stats:datasets}.

\paragraph{The English Dataset} The CovidDialog-English dataset contains 603 English consultations about COVID-19 and other related pneumonia, having 1,232 utterances. The number of tokens (English words) is 90,664.
The average, maximum, and minimum number of utterances in a conversation is 2.0, 17, and 2 respectively. 
The average, maximum, and minimum number of tokens in an utterance is 49.8, 339, and 2 respectively.
Each consultation starts with a short description of the medical conditions of a patient, followed by the conversation between the patient and a doctor. Table \ref{tab:en-data-eg} shows an example. 
The original dialogues are crawled from online healthcare forums, including icliniq.com\footnote{\url{https://www.icliniq.com/en_US/}}, healthcaremagic.com\footnote{\url{https://www.healthcaremagic.com/}}, and healthtap.com\footnote{\url{https://www.healthtap.com/}}. 

\begin{table}[h]
  \centering 
    \caption{Statistics of the English and Chinese dialogue datasets about COVID-19.}
 \begin{tabular}{lll}\hline
 & English & Chinese\\
 \hline
\#dialogs &603& 1,088 \\
\#utterances &1,232& 9,494 \\
\#tokens & 90,664 &  406,550\\
 \hline
 Average \#utterances per dialog &2.0&8.7\\
  Max \#utterances per dialog &17&116\\
   Min \#utterances per dialog &2&2\\
    \hline
 Average \#tokens per utterance &49.8&42.8\\
  Max \#tokens per utterance &339&2,001\\
   Min \#tokens per utterance &2&1\\
\hline
  \end{tabular}
  \label{tb:stats:datasets} 
\end{table}

\paragraph{The Chinese Dataset} The CovidDialog-Chinese dataset contains 1,088 Chinese consultations about COVID-19 and other related pneumonia, having 9,494 utterances. 
In this work, we develop models directly on Chinese characters without performing word segmentation. Each Chinese character in the text is treated as a token. 
The total number of tokens in the dataset is 406,550. The average, maximum, and minimum number of utterances in a conversation is 8.7,  116,  and 2 respectively.  The average,  maximum, and minimum number of tokens in an utterance is 42.8, 2001, and 1 respectively. Each consultation consists of three parts: (1) description of patient's medical condition and history; (2) conversation between patient and doctor; (3) (optional) diagnosis and treatment suggestions given by the doctor. In the description of the patient's medical condition and history, the following fields are included: present disease, detailed description of present disease, what help is needed from the doctor, how long the disease has been, medications, allergies, and past diseases. This description is used as the first utterance from the patient.  The data is crawled from haodf.com\footnote{\url{https://www.haodf.com/}}, which is an online platform of healthcare services, including medical consultation, scheduling appointments, etc. Duplicated and incomplete dialogues were removed.

\begin{table}[h]
  \centering
  \caption{An exemplar consultation in the CovidDialog-English dataset. It consists of a brief description of the patient's medical conditions and the conversation between the patient and a doctor.}
  \small
  \begin{tabular}{p{13.5cm}}\hline
    \textbf{Description of patient's medical condition}: I have a little fever with no history of foreign travel or contact. What is the chance of Covid-19?\\
    \hline
    \textbf{Dialog}\\
    \textbf{Patient}: Hello doctor, I am suffering from coughing, throat infection from last week. At that time fever did not persist and also did not feel any chest pain. Two days later, I consulted with a doctor. He prescribed Cavidur 625, Montek LC, Ambrolite syrup and Betaline gargle solution. Since then throat infection improved and frequent cough also coming out. Coughing also improved remarkably though not completely. From yesterday onwards fever is occuring (maximum 100-degree Celcius). I have not come in touch with any foreign returned person nor went outside. In our state, there is no incidence of Covid-19. Please suggest what to do?\\
   \textbf{Doctor}: Hello, I can understand your concern.
In my opinion, you should get done a chest x-ray and CBC (complete blood count). If both these are normal then no need to worry much. I hope this helps.
\\
 \textbf{Patient}: Thank you doctor. After doing all these I can upload all for further query.\\
  \textbf{Doctor}: Hi, yes, upload in this query only. I will see and revert to you.
\\
    \hline 
  \end{tabular}
  \label{tab:en-data-eg} 
\end{table}

\section{Methods}
In this section, we present several well-established and state-of-the-art methods for dialogue generation. Given a dialogue containing a sequence of alternating utterances between patient and doctor, we process it into a set of pairs $\{(s_i,t_i)\}$ where the target $t_i$ is a response from the doctor and the source $s_i$ is the concatenation of all utterances (from both patient and doctor) before $t_i$. A dialogue generation model takes $s$ as input and generates $t$. The size of the CovidDialog datasets is small. Directly training neural models on these small datasets would result in poor generalization on unseen data. To solve this problem, we utilize transfer learning, which pretrains the neural models on  large corpus, then finetunes the pretrained models on the CovidDialog datasets.

\subsection{Transformer}
Generating response $t$ from the conversation history  $s$ is a typical sequence-to-sequence (seq2seq)~\citep{sutskever2014sequence} modeling problem. Transformer~\citep{vaswani2017attention} is an encoder-decoder architecture for sequence-to-sequence (seq2seq) modeling. Different from seq2seq models~\citep{sutskever2014sequence} that are based on recurrent neural networks (e.g., LSTM~\citep{hochreiter1997long}, GRU~\citep{chung2014empirical}) which model a sequence of tokens via a recurrent manner and hence is computationally inefficient. Transformer eschews recurrent computation and instead uses self-attention which not only can capture the dependency between tokens but also is amenable for parallel computation with high efficiency. Self-attention calculates the correlation among every pair of tokens and uses these correlation scores to create ``attentive" representations by taking weighted summation of tokens' embeddings. Transformer is composed of a stack of building blocks, each consisting of a self-attention layer and a position-wise feed-forward layer.  Residual connection \citep{he2016deep} is applied around each of the two sub-layers, followed by layer normalization~\citep{ba2016layer}. Given the input sequence, an encoder, which is a stack of such building blocks, is applied to obtain a representation for each token. Then the decoder takes these representations as inputs and decodes the sequence of output tokens. To decode the $i$-th token, the decoder first uses self-attention to encode the already decoded sequence $y_1,\cdots,y_{i-1}$, then performs input-output attention between the encodings of $y_1,\cdots,y_{i-1}$ and those of the input sequence. The ``attentive" representations are then fed into a feed-forward layer. The three steps are repeated for multiple times. Finally, the representation is fed into a linear layer to predict the next token. The weight parameters in Transformer is learned by maximizing the conditional likelihood of output sequences conditioned on the corresponding input sequences.

\subsection{GPT}

The GPT model~\citep{radford2018improving} is a language model (LM) based on Transformer. Different from Transformer which defines a conditional probability on an output sequence given an input sequence, GPT defines a marginal probability on a single sequence. Given a sequence of tokens $x_1,\cdots,x_n$, an LM defines a probability on the sequence: $p(x_1,\cdots,x_n)=p(x_1)\prod_{i=2}^n p(x_i|x_1,\cdots,x_{i-1})$, which basically predicts the next token based on the historical sequence. In GPT, $p(x_i|x_1,\cdots,x_{i-1})$ is defined using the Transformer decoder, which first uses a stack of self-attention and feed-forward layers (each followed by layer normalization) to encode $x_1,\cdots,x_{i-1}$, then predicts $x_i$ from the encodings of $x_1,\cdots,x_{i-1}$. 
The weight parameters are learned by maximizing the likelihood on the sequence of tokens.  
GPT-2~\citep{radford2019language} is an extension of GPT, which modifies GPT by moving layer normalization  to the input of each sub-block and adding an
additional layer normalization  after the final self-attention block. Byte pair encoding (BPE) \citep{sennrich2015neural} is used to represent the input sequence of tokens. 

\paragraph{Pretrained GPT models for dialogue generation}
DialoGPT~\citep{zhang2019dialogpt} is a GPT-2 model pretrained on English Reddit dialogues. The dataset is extracted from comment chains in Reddit from 2005 till 2017, comprising 147,116,725
dialogue instances with 1.8 billion tokens. Given a dialogue history $S$ and a ground-truth response $T=x_1,\cdots,x_n$, DialoGPT is trained to maximize the following probability:  $p(T|S)=p(x_1|S)\prod_{i=2}^{n}p(x_i|S,$\\$x_1,\cdots,x_{i-1})$, where conditional probabilities are defined by the Transformer decoder. A maximum mutual information (MMI)~\citep{li2015diversity} scoring function is used to penalize generated responses that are bland. We finetune DialoGPT on the CovidDialog-English dataset for generating English COVID-19 dialogues. GPT2-chitchat\footnote{\url{https://github.com/yangjianxin1/GPT2-chitchat}} is a GPT-2 model pretrained on Chinese Chatbot Corpus\footnote{\url{https://github.com/codemayq/chinese_chatbot_corpus}} which contains about 14M dialogues and 500k-Chinese-Dialog\footnote{\url{https://drive.google.com/file/d/1nEuew_KNpTMbyy7BO4c8bXMXN351RCPp/view}} which contains 500K Chinese dialogues. The training strategy of GPT2-chitchat is the same as that of DialoGPT. We finetune GPT2-chitchat on our CovidDialog-Chinese dataset for generating Chinese COVID-19 dialogues.

\subsection{BERT-GPT}
BERT-GPT~\citep{wu2019importance} is a model used for dialogue generation where pretrained BERT~\citep{devlin2018bert} is used to encode the conversation history and GPT is used to generate the responses. While GPT focuses on learning a Transformer decoder for text generation purposes, BERT~\citep{devlin2018bert} aims to  learn a Transformer encoder for representing texts. BERT’s model architecture is a multi-layer bidirectional Transformer encoder. In BERT, the Transformer uses bidirectional self-attention, whereas in GPT every
token can only attend to context to its left. To train the encoder, BERT masks some percentage of the input
tokens at random, and then predicts  those masked tokens by feeding the final hidden vectors (produced by the encoder) corresponding to the mask tokens into an output softmax over
the vocabulary. Since BERT leverages context to both the left and the right for representing a token, it presumably has better representation power than GPT which only leverages context to the left. In dialogue generation, for the given conversation history, instead of using GPT for obtaining the representation, we can use a more powerful pretrained BERT to encode it. The BERT encoding of the conversation history is fed into GPT to generate the response. 

In BERT-GPT, the pretraining of the BERT encoder and the GPT decoder is conducted separately, which may lead to inferior performance.  Auto-Regressive
Transformers (BART)~\citep{lewis2019bart} has a similar architecture as BERT-GPT, but pretrains the encoder and decoder jointly. To pretrain the BART weights, the input text is corrupted randomly, such as token masking, token deletion, text infilling, etc., then the network is
learned to reconstruct the original text. BART is pretrained on the data used in \citep{liu2019roberta}, consisting of 160Gb of news, books, stories, and web texts.

\paragraph{Pretrained BERT-GPT models for dialogue generation}

BERT-GPT-Chinese~\citep{wu2019importance} is a BERT-GPT model pretrained on Chinese corpus. For the BERT encoder in BERT-GPT-Chinese, it is set to the Chinese BERT~\citep{cui2019pre}, which is a large-scale pretrained BERT model on Chinese texts. For the GPT decoder in BERT-GPT-Chinese, it has the same architecture as BERT but applies lower-triangular mask for autoregressive text generation. The decoder is initialized  with Chinese BERT’s weights. Then the decoder is pretrained with a maximum likelihood estimation (MLE) objective on a large-scale
multi-domain Chinese corpus. The resulting model consists of a bidirectional Transformer as the encoder, a unidirectional Transformer as the decoder, and an attention mechanism to connect them. The Chinese corpus used for pretraining is collected from the Large Scale Chinese Corpus for NLP\footnote{\url{https://github.com/brightmart/nlp_chinese_corpus}}, including the following datasets: Chinese Wikipedia which contains 104M articles, News which contains 2.5 million news articles from 63,000 sources, Baike QA which is a wiki question
answering (QA) dataset with 1.5 million QA pairs from 493 different domains, and Community QA which contains 4.1 million
comments and 28 thousand topics. The total size of these datasets is 15.4 GB. We finetune BERT-GPT-Chinese on the CovidDialog-Chinese dataset for Chinese COVID-19 dialogue generation. For English COVID-19 dialogue generation, we finetune the pretrained BART model on the CovidDialog-English dataset.

\section{Experiments}

\subsection{Experiments on the English Dataset}
\subsubsection{Experimental Settings}
\begin{table}[h]
    \centering
        \caption{English dataset split statistics}
    \begin{tabular}{@{}lccc@{}}
    \toprule
    Split      & \#Dialogues & \# Utterances & \#  Pairs \\ \midrule
    Train      & 482         & 981           & 490                       \\
    Validation & 60         & 126           & 63                       \\
    Test       & 61         & 122           & 61                        \\
    \bottomrule
    \end{tabular}
    \label{tab:data_split}
\end{table}
For the English dataset, we split it into a training, a validation, and a test set based on dialogues, with a ratio of 8:1:1. Table~\ref{tab:data_split} shows the statistics of the data split. 
The hyperparameters were tuned on the validation dataset. For all methods, we used the Adam~\citep{kingma2014adam} optimizer with linear learning rate scheduling, setting the initial learning rate as 4e-5 and the batch size as 4. The objective is the cross entropy loss with label smoothing where the factor was set to 0.1. For pretrained models, we finetune them on the CovidDialog-English dataset for 5 epochs, while for the un-pretrained Transformer, we train it for 50 epochs. We set a checkpoint at the end of every epoch and finally take the one with the lowest perplexity on validation set as the final model. In response generation, for all models, we use beam search with beam width of 10 as our decoding strategy. For DialoGPT~\citep{zhang2019dialogpt}, we used three variants with different sizes: DialoGPT-small, DialoGPT-medium, DialoGPT-large, with 117M, 345M and 762M weight parameters respectively. Maximum mutual information was not used.


\begin{table}[h]
    \centering
        \caption{Performance on the CovidDialog-English test set.}
    \begin{tabular}{@{}llllll@{}}
\toprule
           &\multirow{2}{*}{Transformer}  & \multicolumn{3}{c}{DialoGPT}&
            \multirow{2}{*}{BART}   \\
            \cline{3-5}
           & &Small &Medium & Large
           &\\
           \midrule
Perplexity & 263.1      & 28.3         & 17.5           & 18.9         & \tb{15.3}  \\
NIST-4     & 0.71     & 1.90           & 2.01            & \tb{2.29}           & 1.88   \\
BLEU-2     & 7.3\%      & 9.6\%       & 9.4\%
& \tb{11.5}\%         & 8.9\% \\
BLEU-4     & 5.2\%      & 6.1\%         & 6.0\%          & \tb{7.6}\%         & 6.0\% \\
METEOR     & 5.6\%      & 9.0\%         & 9.5\%          & \tb{11.0}\%         & 10.3\% \\
Entropy-4  & 5.0      & 6.0        & \tb{6.6}          & \tb{6.6}         & 6.5 \\
Dist-1     & 3.7\%      & 9.5\%         & 16.6\%          & 13.9\%         & \tb{16.8}\% \\
Dist-2     & 6.4\%      & 22.9\%         & \tb{36.7}\%          & 31.0\%         & 35.7\% \\
Avg. Len   & 40.0       & 51.3          & 50.1           & 54.4          & 45.4  \\ \bottomrule
\end{tabular}
    \label{tab:results-en}
\end{table}

We performed automatic evaluation, using metrics including perplexity, NIST-$n$~\citep{doddington2002automatic} (where $n=4$), BLEU-$n$~\citep{papineni2002bleu} (where  $n=2$ and 4), METEOR~\citep{lavie2007meteor}, Entropy-$n$~\citep{zhang2018generating} (where $n=4$), and Dist-$n$~\citep{li2015diversity} (where  $n=1$ and 2). BLEU, METEOR, and NIST are common metrics for evaluating machine translation. They compare the similarity between generated responses and the ground-truth by matching $n$-grams. NIST is a variant of BLEU, which weights $n$-gram
matches using information gain to  penalize uninformative $n$-grams. Perplexity is used to measure the quality and smoothness of generated responses. Entropy and Dist are used to measure the lexical diversity of generated responses. For perplexity, the lower, the better. For other metrics, the higher, the better. 

As noted in~\citep{liu2016not}, while automatic evaluation is useful, they are not completely reliable. To address this issue, we perform human evaluation of the generated responses. Five undergraduate and graduate students are asked to give ratings (from 1 to 5, higher is better) to responses in three aspects: (1) Relevance: how relevant the response is to the conversation history; (2) Informativeness: How much medical information and suggestions are given in the response; (3) Doctor-like: How the response sounds like a real doctor. The responses are de-identified: annotators do not know a response is generated by which method. The groundtruth response from the doctor is also given ratings (in an anonymous way). Human evaluation was conducted on the test examples in the CovidDialog-English dataset. The ratings from different annotators are averaged.

\subsubsection{Results}

\begin{table}[h]
    \centering
        \caption{Human evaluation on the CovidDialog-English test set.}
    \begin{tabular}{@{}lcccc@{}}
\toprule
           &\multirow{2}{*}{Transformer}  & DialoGPT&
            \multirow{2}{*}{BART} &  Ground  \\
         &  & Large& & truth\\
           \midrule
Relevance & 2.45 & 2.98 & 3.04 & 3.59\\
Informativeness & 2.66 & 2.60 & 2.77 & 3.53 \\
Doctor-like & 2.32 & 3.20 & 3.36 & 3.50 \\
\bottomrule
\end{tabular}
    \label{tab:humanresults-en}
\end{table}

Table~\ref{tab:results-en} summarizes the automatic evaluation results achieved by different methods. From this table, we make the following observations. First, pretrained models including DialoGPT and BART in general perform better than un-pretrained Transformer. This demonstrates the effectiveness of transfer learning, which leverages external large-scale data to learn powerful representations of texts. Second, BART achieves lower perplexity than DialoGPT models. This is probably because BART is pretrained on a much larger and more diverse corpus than DialoGPT, which enables BART to better model the language. Third, DialoGPT-large performs better than BART on machine translation metrics including NIST, BLEU, and METEOR. This is probably because DialoGPT-large is pretrained on dialogue data and therefore tends to generate $n$-grams that are more related to dialogues. Fourth, on diversity-related metrics including Entropy and Dist, BART is on par with DialoGPT models. 
The average length of the generated responses by different methods is close to that of the ground-truth, which is around 50.

\begin{table}[h!]
    \caption{Generated responses on a test example in the CovidDialog-English dataset. }
  \centering
  \small
  \begin{tabular}{p{13.5cm}}\hline
    \textbf{Patient}: i have all the symptoms except fever, i went to medicross and dr said i can get tested if i want to i'm not sure if i should. she gave me antibiotics klacid xl 500mg, she said i can take it if i feel worse i'm worried it will make immune system bad?
\\
    \hline
    \textbf{Groundtruth response}: in brief: antibiotic i don't recommend antibiotics for a simple viral upper respiratory tract infection unless examination revealed signs of acute bronchitis or sinusitis. they are not effective for viral infections like covid 19 with no bacterial lung involvement either. if you've been exposed to someone with covid 19 or or if you or someone you were exposed to travelled to a region where it was endemic, get tested would you like to video or text chat with me?
\\
    \hline \textbf{Transformer}: hello and welcome to ‘ask a doctor’ service.i have reviewed your query and here is my advice. i have reviewed your query. let me know if i can assist you further.
\\
    \hline \textbf{DialoGPT-small}: in brief: yes. if you feel worse you should go to a doctor. would you like to video or text chat with me? let me know if i can assist you with any further questions. regards dr.jolanda jolanda.
 \\
\hline \textbf{DialoGPT-medium}:  in brief: fever, dry cough, shortness of breath, fever can be from strep throat (antibiotics for strep), or from a strep infection, or from some other cause such as allergies, irritants, or irritants. would you like to video or text chat with me?
\\
\hline \textbf{DialoGPT-large}: in brief: do you have access to a telephonic consult, or do you need to go in person? would you like to video or text chat with me? i can understand your concern. i have reviewed your query and here is my advice. yes, telephonically would be best. would you be happy to chat with you? i have answered your query. let me know if i can assist you further. regards, dr jnikolla, pulmonologist
\\
\hline \textbf{BART}:  hello and welcome to ‘ask a doctor’ service. i have reviewed your query and here is my advice. yes, you should get tested for covid-19. get tested if you have fever, cough or shortness of breath. if you are a smoker or have been in contact with someone with covid, get tested. would you like to video or text chat with me?
\\
\hline
  \end{tabular}
  \label{tab:en-eg1} 
\end{table}

Table~\ref{tab:humanresults-en} shows the human evaluation results. From this table, we make the following observations. 
\begin{table}[h]
  \centering 
    \caption{Chinese dataset split statistics}
 \begin{tabular}{llll}\hline
 Split & \# Dialogues& \# Utterances& \# Pairs\\
 \hline
Train & 870 &7844&3922 \\
Validation& 109  &734   &367 \\
Test &109  & 916  &458 \\
\hline
  \end{tabular}
  \label{tb:stats:datasplit:chinese}
\end{table} 
First, pretrained models including DialoGPT and BART perform better than Transformer. This further demonstrates the effectiveness of pretraining. Second, BART performs better than DialoGPT, though DialoGPT achieves better scores on machine translation metrics. This is in accordance with the results in \citep{liu2016not} that machine translation metrics are not good for evaluating dialogue generation.  Third, BART achieves a doctor-like score that is close to the groundtruth. This indicates that the auto-generated responses have high language quality. The relevance rating of BART is higher than 3, which indicates a good level of relevance between the generated responses and conversation histories. BART's informativeness rating is better than Transformer and DialoGPT, but has a large gap with that of the groundtruth. Additional efforts are needed to improve informativeness, such as incorporating medical knowledge.

\begin{table}[h]
  \centering 
    \caption{Performance on the CovidDialog-Chinese test set.}
  \begin{tabular}{lllll}\hline
  &\multirow{2}{*}{Transformer}  & \multicolumn{2}{c}{DialoGPT}&
            \multirow{2}{*}{BERT-GPT}   \\
            \cline{3-4}
           & &No MMI &MMI &\\
 \hline
 Perplexity &  53.3& 22.1& 25.7 &\tb{10.8}\\
  NIST-4 &0.39 & 0.43 & \tb{0.46} &0.36\\
  BLEU-2& 5.7\%& 6.2\% & \tb{7.2}\% &4.6\%\\
  BLEU-4& 4.0\%& 4.0\% & \tb{5.4}\% &2.8\%\\
  METEOR &13.5\% & 13.9\%& \tb{14.3}\% &12.2\%\\
  Entropy-4 &7.9 & 9.0& \tb{9.1} &8.5\\
  Dist-1& 5.5\%& 5.9\% & 3.2\% &\tb{7.9}\%\\
  Dist-2& 29.0\%& 38.7\% & 35.7\% &\tb{39.5}\%\\
  Avg Len & 19.3 & 35.0& 58.7 &21.6\\
   \hline
  \end{tabular}
  \label{tb:res:chinese} 
\end{table}

Table~\ref{tab:en-eg1} shows an example of generating a doctor's response given the utterance of a patient. As can be seen, the response generated by BART is more relevant, informative, and human-like, compared with those generated by other baselines.  BART's response suggests the patient to get tested for COVID-19 since the patient stated that ``I have all the symptoms except fever". This response gives correct and informative medical advice: ``get tested if you have fever, cough, or shortness of breath", ``if you are a smoker or have been in contact with someone with covid, get tested". The response is human-like, with correct grammar and semantics. It begins with a welcome opening, then provides medical advice, and finally offers to further discuss via video. In contrast, the response generated by DialoGPT-large is not informative. It does not provide any useful medical advice. The response generated by DialoGPT-medium is informative, but not very relevant. The patient has no fever, but this response focuses on talking about the causes of fever. Similar to DialoGPT-large, the responses generated by DialoGPT-small and Transformer are uninformative.

\begin{table}[h]
    \centering
        \caption{Human evaluation on the CovidDialog-Chinese test set.}
    \small
    \begin{tabular}{@{}lcccc@{}}
\toprule
           &\multirow{2}{*}{Transformer}  & DialoGPT&
            \multirow{2}{*}{BERT-GPT}  &
            \multirow{2}{*}{Groundtruth}\\
         &  & No MMI& \\
           \midrule
Relevance &2.24 & 1.82 & 2.65& 3.42\\
Informativeness &2.06 & 1.72 &2.37 &3.26\\
Doctor-like & 2.57& 1.80 & 3.16& 3.78\\
\bottomrule
\end{tabular}
    \label{tab:humanresults-cn}
\end{table}

\subsection{Experiments on the Chinese Dataset}
\subsubsection{Experimental settings}

Based on dialogues, we split the Chinese dataset into a training set, validation set, and test set, with a ratio of 8:1:1. 
Table~\ref{tb:stats:datasplit:chinese} shows the statistics of the data split.  
The hyperparameters were tuned on the validation set. We stop the training procedure when the validation loss stops to decrease.  For DialoGPT, we used the DialoGPT-small architecture where the number of layers in the Transformer was set to 10. The context size was set to 300. The embedding size was set to 768. The number of heads in multi-head self-attention was set to 12. The epsilon parameter in layer normalization was set to 1e-5. Network weights were optimized with Adam, with an initial learning rate of 1.5e-4 and a batch size of 8. The Noam learning rate scheduler with 2000 warm-up steps was used. 
In the finetuning of BERT-GPT, the max length of the source sequence and target sequence was set to 400. The encoder and decoder structures are similar to those in BERT, which is a Transformer with 12 layers and the size of the hidden states is 768. The network weights are optimized with stochastic gradient descent with a learning rate of 1e-4. For Transformer, we used the HuggingFace implementation\footnote{https://github.com/huggingface/transformers} and followed their default hyperparameter settings. During decoding for all methods, beam search with $k=50$ was used. We evaluated the models using perplexity, NIST-$4$, BLEU-$2,4$, METEOR, Entropy-$4$, and Dist-$1,2$. Human evaluation was conducted by 5 graduate and undergraduate students, on 100 randomly-sampled examples from the test set of CovidDialog-Chinese. The ratings from different annotators are averaged.

\subsection{Results on the Chinese Dataset}
Table~\ref{tb:res:chinese} summarizes the automatic evaluation results. From this table, we make the following observations. First, pretrained models including DialoGPT and BERT-GPT achieve lower perplexity than Transformer. This further demonstrates the effectiveness of transfer learning. Second, DialoGPT-MMI achieves better scores on machine translation metrics, which is consistent with the results on the CovidDialog-English dataset. Third, BERT-GPT achieves better Dist scores than other methods. We manually checked the generated responses by BERT-GPT. Indeed, they are more diverse than others. Fourth, maximum mutual information (MMI) does not have a clear efficacy in improving the quality of generated responses.

Table~\ref{tab:humanresults-cn} shows the human evaluation results. As can be seen, pretrained BERT-GPT works better than unpretrained Transformer. Though pretrained, DialoGPT is not as good as Transformer. The possible reason is the training corpora of DialoGPT is daily dialogues, which has a large domain shift from medical dialogues. The performance gap between BERT-GPT and Groundtruth is larger than that between BART and Groundtruth, despite the number of Chinese training dialogues is larger than that of English training dialogues. This indicates that it is more challenging to develop COVID-19 dialogue systems on Chinese. One major reason is the Chinese dialogues are more noisy than the English ones, with a lot of incorrect grammars, abbreviations, semantic ambiguities, etc.

Figure 1 shows an example of generating a doctor's response given the utterance of a patient. The response generated by BERT-GPT tells the patient that it is not likely to be COVID-19. This is a reasonable response since the patient mentioned that he/she was tested negative. The response generated by DialoGPT is not understandable. The response generated by Transformer is ambiguous. There is a comma between ``No" and ``COVID-19". It is difficult to judge whether the response is suggesting ``having COVID-19" or ``having no COVID-19".

\begin{figure}[h!]
\begin{center}
\caption{Generated responses on a test example in the CovidDialog-Chinese dataset.}
\includegraphics[width=\textwidth]{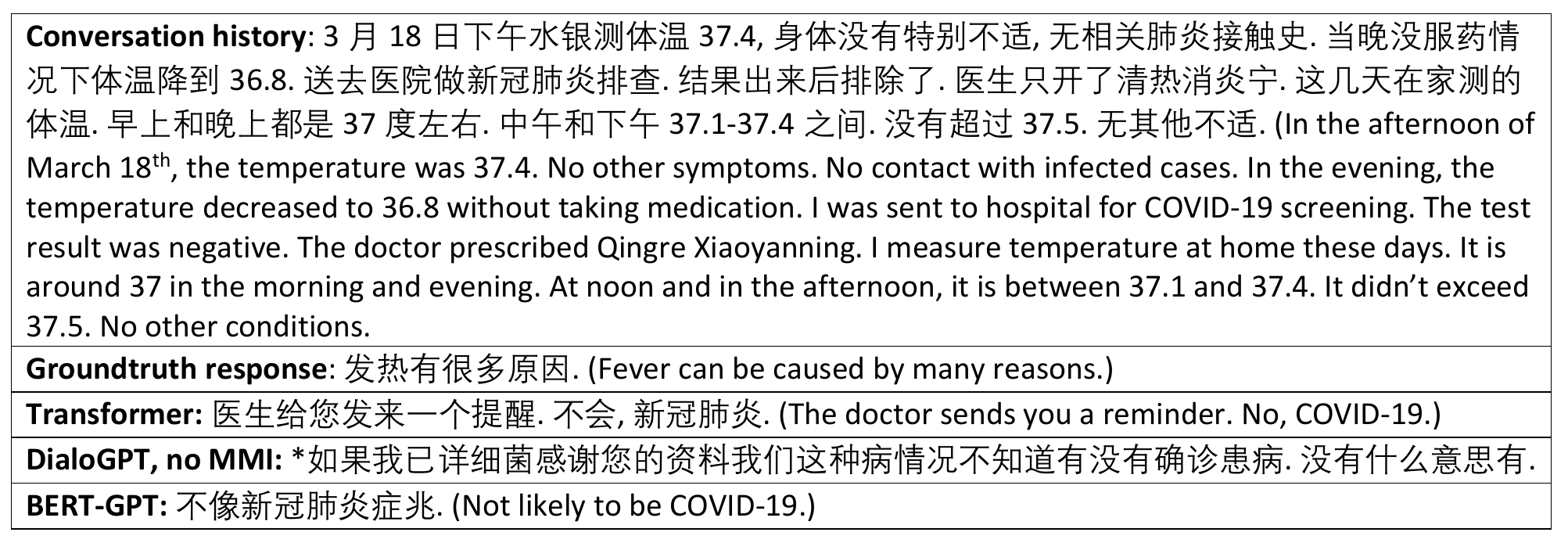}
\end{center}
\label{res-eg-cn}
\end{figure}

\section{Related Works}
Many works have been devoted to developing medical dialogue systems. Please refer to \citep{laranjo2018conversational} for a comprehensive review. Some methods~\citep{lucas2017reporting,philip2017virtual,tanaka2017embodied} predefine a sequence of steps or states which are used to guide the conversation. Other methods~\citep{rhee2014mobile,ireland2016hello,fitzpatrick2017delivering} use predetermined templates to extract information from the conversation history and use rules to generate responses from the filled slots in the templates. These methods rely heavily on knowledge engineering and are difficult to be quickly adapted to a new and time-sensitive task such as COVID-19 dialogue generation.

Data-driven medical dialogue generation based on neural networks has been investigated in several works.  Wei et al.~\citep{wei2018task} proposed a task-oriented dialogue system to
make medical diagnosis automatically based on reinforcement learning. The system  converses with patients to collect additional symptoms beyond their self-reports. Xu et al.~\citep{xu2019end} proposed a knowledge-routed relational dialogue system  that
 incorporates  medical knowledge graph into  topic transition in dialogue management. Xia et al.~\citep{xiagenerative} developed a reinforcement learning (RL) based dialogue system for automatic diagnosis. They proposed a policy gradient framework based on the generative adversarial network  to optimize the RL model. In these works, the neural models are trained from scratch on small-sized medical dialogue datasets, which are prone to overfitting.

\section{Conclusions}
In this work, we make the first attempt to develop dialogue systems to provide medical consultations about COVID-19. To achieve this goal, we first collected two datasets -- CovidDialogs -- which contain medical conversations between patients and doctors about COVID-19. Then on these datasets, we train dialogue generation models based on  Transformer, DialoGPT, and BERT-GPT pretrained on large-scale dialogue datasets and other corpus. Human evaluation and automatic evaluation results show that these models are promising in generating clinically meaningful and linguistically high-quality consultations for COVID-19.


\bibliography{release}

\end{document}